\documentclass[conference]{IEEEtran}
\IEEEoverridecommandlockouts

\usepackage{cite}
\usepackage{booktabs}
\usepackage{array}
\usepackage{amsmath,amssymb,amsfonts}
\usepackage{algorithmic}
\usepackage{graphicx}
\usepackage{textcomp}
\usepackage{tikzpagenodes}
\usepackage{eso-pic}
\usepackage[colorlinks=true, linkcolor=blue, citecolor=blue]{hyperref}
\usepackage[all]{hypcap}

\hypersetup{
    colorlinks=true,
    linkcolor=blue,
    linktocpage=true,
    pdfdisplaydoctitle=true,
    bookmarksnumbered=true,
    pdfstartview={XYZ null null null}
}

\def\BibTeX{{\rm B\kern-.05em{\sc i\kern-.025em b}\kern-.08em
    T\kern-.1667em\lower.7ex\hbox{E}\kern-.125emX}}

\begin{document}

\AddToShipoutPictureFG*{%
  \begin{tikzpicture}[remember picture,overlay]
    \node[anchor=south west, yshift=-40pt]
      at (current page text area.south west) {%
        \normalsize 979-8-3195-0680-1/26/\$31.00~\copyright2026~IEEE%
      };
  \end{tikzpicture}%
}

\title{ORB-SVM: An Innovative Hybrid Framework for Efficient Brain Tumor Detection from MRI Scans}

\author{
\IEEEauthorblockN{Amirhosein Azarpour$^{*}$}
\IEEEauthorblockA{\textit{Department of Computer Science} \\
\textit{Shahid Beheshti University} \\
Tehran, Iran \\
amirazarpour07@gmail.com}
}

\maketitle

\begin{abstract}
Brain cancer remains one of the most significant challenges in modern medicine, where the accuracy of early stage diagnosis is a decisive factor in patient survival and treatment efficacy. Although Magnetic Resonance Imaging (MRI) is the established gold standard for visualizing neurological structures, the interpretation of these high dimensional scans is often complicated by subjective variability among practitioners and the inherent noise present in complex medical images. While contemporary approaches frequently rely on high parameter deep learning architectures, such models often involve significant computational costs and require extensive data for effective training. This study introduces a hybrid framework that utilizes the Oriented FAST and Rotated BRIEF (ORB) algorithm for precise feature extraction and a Support Vector Machine (SVM) for classification~\cite{Rublee2011,Hearst1998}. The proposed approach achieves a substantial data reduction of approximately 99.5\%, which effectively minimizes the influence of non informative background data while preserving critical diagnostic patterns essential for tumor identification. By balancing feature sparsity with a robust kernel based classifier, this methodology addresses the limitations of over parameterized systems while maintaining high diagnostic integrity. Experimental evaluations conducted on the Br35H dataset demonstrate that the framework attains a classification accuracy of 97.5\%. The findings suggest that the integration of localized feature representation and optimized classification provides a reliable and resource efficient alternative for medical image analysis, offering a structured solution that maintains performance without the need for extensive computational overhead.
\end{abstract}


\section{Introduction}

Brain neoplasms represent a persistent and formidable challenge within modern oncology, where diagnostic precision and timely intervention are decisive factors in patient prognosis. This research proposes a robust diagnostic methodology that integrates Oriented FAST and Rotated BRIEF (ORB) for feature extraction with a Support Vector Machine (SVM) for classification~\cite{Shamla2023}. Beyond its technical framework, the significance of this approach lies in its potential to mitigate the inherent inconsistencies associated with manual radiological interpretation. By automating the feature identification process, the proposed system aims to reduce the subjective variability often encountered during the clinical evaluation of complex MRI scans.

The implications of this framework extend to the broader field of medical image analysis, offering a reliable alternative to traditional diagnostic workflows. The automation of both feature extraction and subsequent classification enhances the consistency of the results, thereby minimizing the reliance on subjective judgments and reducing the ambiguity often associated with pathological patterns in neuroimaging. Furthermore, the modular architecture of this methodology suggests potential adaptability to other oncological domains or neurological disorders, provided the inherent variability of medical imaging datasets is carefully addressed.

The synergy of the selected feature representation and classification techniques achieves high accuracy while maintaining operational efficiency, thus contributing to the ongoing integration of machine learning in clinical environments. This research positions itself as a balanced contribution to a field often characterized by the tension between cautious skepticism and rapid technological adoption. Future developments involve the expansion of training datasets and the potential for clinical integration to validate the model's efficacy in real-world diagnostic settings. While the application of this specific hybrid framework to MRI based tumor detection represents a significant advancement, a measured perspective is maintained regarding its long term maturation in complex clinical infrastructures.In contrast to high parameter architectures such as Convolutional Neural Networks (CNNs), which typically require intensive pixel-level analysis, the proposed methodology adopts a selective feature oriented strategy~\cite{Vianna2018}. While deep learning models often necessitate substantial computational resources due to the high dimensionality of MRI data, the current approach focuses on extracting only the most salient descriptors. This reduction in the computational burden ensures that the system remains resource efficient and suitable for the operational constraints of clinical environments, where diagnostic processing must be both reliable and timely~\cite{Sazal2015,Afshar2018}.
Experimental validation on a diverse MRI dataset yielded a classification accuracy of 97.5\%. Given the global prevalence of brain tumors—where meningiomas alone constitute approximately one-third of diagnosed cases—the necessity for scalable and accurate diagnostic tools is evident. 
Magnetic Resonance Imaging (MRI) remains the fundamental modality for neurological imaging due to its superior soft tissue contrast. The technology operates on the principle of nuclear magnetic resonance, where radiofrequency pulses excite protons within the body's tissues. Upon the cessation of the pulse, the energy released as protons return to their equilibrium state is measured. Most of these signals are derived from hydrogen protons, which are abundant in water and fatty tissues; their behavior within a magnetic field provides the data necessary to visualize the brain's internal structures.

\begin{figure}[htbp]
    \centering
    \includegraphics[width=0.5\textwidth]{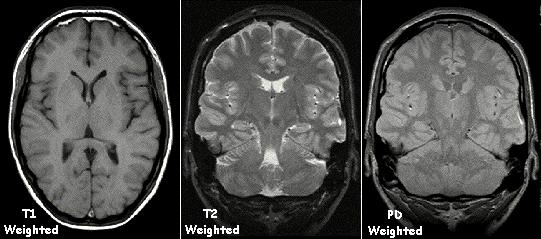}
    \caption{Examples of T1 weighted, T2 weighted and PD weighted MRI scans. Adapted with permission by Kieran Maher using Graphic Converter.}
    \label{1}
\end{figure}

The diagnostic process is methodical: the patient is subjected to a uniform magnetic field, typically 1.5 Tesla, causing protons to align and precess. A radiofrequency pulse subsequently disrupts this alignment, and the signals captured during the relaxation phase are processed. The variations in the rates at which different tissues return to equilibrium generate the contrast observed in the final images (Fig. \ref{1}). This visual narrative is primarily shaped by T1 and T2 relaxation processes, which determine the brightness and texture of the tumorous tissue relative to the surrounding healthy anatomy. Spatial encoding, achieved through magnetic gradients, enables the localization of these signals, culminating in a 2D image through Fourier transformations.

\section{Related Works}
Traditional machine learning methodologies have long established a robust foundation in medical image classification, with Support Vector Machines (SVM) serving as a prominent benchmark for supervised learning tasks~\cite{Hearst1998}. Research by Paredes et al., for instance, utilized localized image patches integrated with k-nearest neighbor classification, demonstrating significant accuracy in early diagnostic models~\cite{Paredes2002}. Subsequently, Parveen and Sathik employed a distinct methodology by combining discrete wavelet transforms with fuzzy C-means clustering for pneumonia detection in thoracic radiographs~\cite{Parveen2011}. Furthermore, Caicedo et al. integrated Scale-Invariant Feature Transform (SIFT) descriptors with SVM classifiers, reporting highly precise results~\cite{Caicedo2009}. While these classical approaches significantly advanced the field, they were characterized by a heavy reliance on manual feature engineering and occasionally exhibited limited robustness when faced with variations in image quality or multi scale inconsistencies~\cite{Paredes2002,Parveen2011,Caicedo2009}.

Chandrasekhara et al. proposed a SIFT-SVM framework for prostate cancer detection in MRI scans, extracting SIFT features and performing classification through a multiclass SVM model~\cite{Chandrasekhara2020}. The results validated SIFT's resilience against affine transformations and confirmed its compatibility with SVM in addressing complex classification challenges~\cite{Chandrasekhara2020}.

Yadav and Jadhav evaluated three distinct strategies for pneumonia detection in chest X-rays: linear SVM with ORB features, transfer learning via VGG16 and InceptionV3, and capsule networks~\cite{Yadav2019}. Their analysis indicated a performance advantage for transfer learning, particularly in scenarios with limited datasets, where convolutional neural networks (CNNs) identified more intricate patterns compared to traditional descriptors~\cite{Yadav2019}. Similarly, Kermany et al. achieved competitive accuracy in classifying optical coherence tomography images through InceptionV3 transfer learning, demonstrating performance levels comparable to those of experienced clinical experts~\cite{Kermany2018}.

\section{Material and Methods}

\subsection{Dataset}\label{AA}

The dataset for this work consists of 3000 MRI scans, including two classes of healthy and cancerous brains. These scans originate from hospital patients and cover a range of MRI modalities.The dataset employed in this research is Br35H: Brain Tumor Detection 2020, which is publicly available on the Kaggle platform\cite{Br35H}.

\subsection{Preprocessing}
All MRI images are preprocessed prior to feature extraction. Images are resized to $256\times 256$ pixels and normalized to $[0,1]$ range to ensure consistency across different scanners. The dataset of 3000 scans is stratified to maintain class balance and split into training ($80\%$, 2400 images) test ($20\%$, 600 images) subsets.

\subsection{ORB Algorithm Principle (Orient Fast and Rotated Brief)}
Oriented Fast and Rotated Brief (ORB) is based on the famous fast feature detection and brief feature descriptor~\cite{Xie2022,Chang2025,Rublee2011}.

\subsection{Feature Point Detection}
Feature points in an image mark locations that carry meaningful visual information places like edges, bright spots in otherwise dark areas, or dark spots in bright regions. The ORB algorithm tackles the problem of identifying these points quickly, using an efficient detection strategy. At the heart of ORB is the FAST method (Features from Accelerated Segment Test), which looks for pixels that noticeably differ from their neighbors. Essentially, a pixel is considered a feature point if its intensity stands out compared to most of the surrounding pixels, as formulated in \eqref{eq:fast_n}:

\begin{equation} \label{eq:fast_n}
N = \sum_{x \in \operatorname{circle}(p)} |I(x) - I(p)| > \varepsilon_d
\end{equation}

In this framework, $I(x)$ denotes the grayscale intensity of a point along the circumference surrounding the candidate pixel $p$, $I(p)$ represents the intensity of the central pixel, and $\varepsilon_d$ is the intensity difference threshold. A pixel $p$ is considered a feature point if the number of surrounding pixels exceeding this threshold, $N$, surpasses a predefined proportion, typically three quarters of the circle.

\begin{figure}[htbp]
    \centering
    \includegraphics[width=0.5\textwidth]{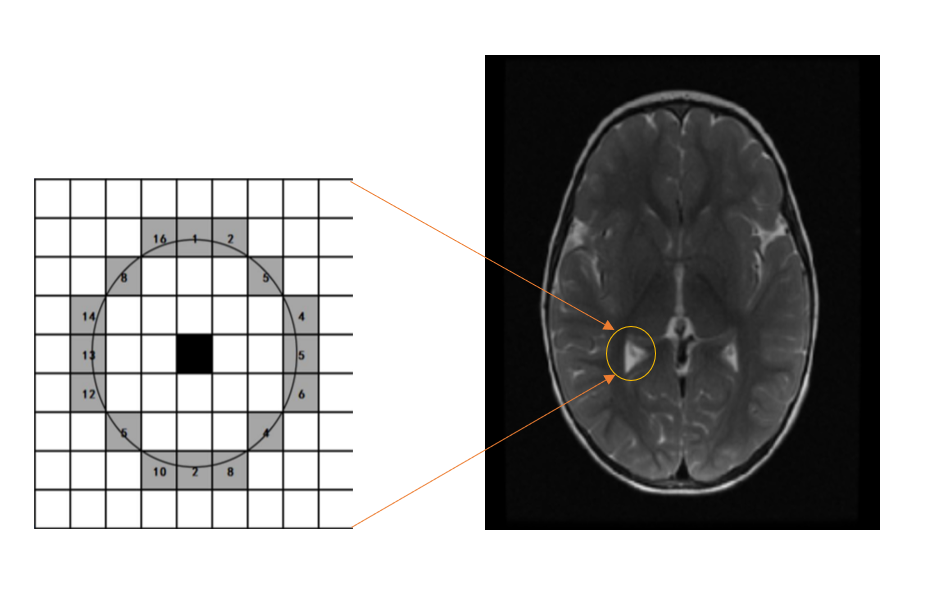}
    \caption{Schematic diagram of the extraction of FAST feature points.}
    \label{2}
\end{figure}

The FAST detection procedure involves selecting a candidate pixel $P$ with intensity $M$ and defining a threshold $V$, often set to 20\% of $M$, to assess significant intensity differences. Sixteen pixels are sampled along a circle of radius three centered on $P$. The candidate is identified as a corner if at least $L$ consecutive pixels on the circle are either all brighter than $M + V$ or all darker than $M - V$. Commonly, $L$ is set to 12, meaning that the candidate is accepted as a feature point if this condition is satisfied; otherwise, it is rejected. The visual representation of this extraction process is shown in Fig. \ref{2}.

To enhance computational efficiency, a preliminary test is applied: four pixels at 90 degree intervals around the candidate are first examined. If at least three of these pixels display sufficient intensity difference from the center, the complete evaluation proceeds; otherwise, the candidate is discarded.

Since the FAST algorithm does not provide orientation information for feature points, the ORB algorithm introduces two strategies to enhance invariance. First, a three layer Gaussian image pyramid is constructed to improve scale invariance. Second, the gray centroid method is applied to assign a rotation angle to each feature point. Specifically, a coordinate system is established with the feature point as the origin, and the centroid of the neighborhood $S$ is calculated.

The moments of the neighborhood $S$ are computed as shown in \eqref{eq:moments}:
\begin{equation} \label{eq:moments}
M_{pq} = \sum_{x, y} x^p y^q I(x, y),
\end{equation}
where $I(x, y)$ represents the grayscale intensity of the image, and $x, y \in [-r, r]$, with $r$ denoting the radius of the feature point neighborhood. 

The centroid $C$ of the neighborhood is then derived using these moments, as given by \eqref{eq:centroid}:
\begin{equation} \label{eq:centroid}
C = \left( \frac{M_{10}}{M_{00}}, \frac{M_{01}}{M_{00}} \right).
\end{equation}

The orientation $\theta$ of the feature point is finally calculated as formulated in \eqref{eq:orientation}:
\begin{equation} \label{eq:orientation}
\theta = \arctan \left( \frac{M_{01}}{M_{10}} \right).
\end{equation}

To ensure rotation invariance, it is essential that the coordinates $x$ and $y$ are confined within the circular neighborhood of radius $r$, i.e., $x, y \in [-r, r]$~\cite{Xie2022}.

\subsection{Calculate Feature Point Descriptors}
ORB utilizes the improved BRIEF algorithm to compute the descriptor of a feature point, addressing the primary limitation of BRIEF, namely its lack of rotation invariance. The core idea is to select $N$ point pairs in a specific pattern around a feature point $P$ and combine the comparison results of these $N$ pairs to form the descriptor.

The BRIEF descriptor is based on the assumption that a small image neighborhood can be represented using relatively few intensity comparisons. For an $S \times S$ image patch $P$, the binary test $\tau$ is defined in \eqref{eq:binary_test} as:

\begin{equation} \label{eq:binary_test}
\tau(p; x, y) =
\begin{cases}
1, & \text{if } p(x) < p(y), \\
0, & \text{otherwise},
\end{cases}
\end{equation}

where $p(x)$ is the pixel intensity at location $x = (u,v)^T$ in the image patch centered at $P$. Selecting $N_d$ such point pairs $(x, y)$ uniquely defines the binary criterion. The BRIEF descriptor is thus a binary string of $N_d$ bits, which is formulated in \eqref{eq:brief_descriptor}:

\begin{equation} \label{eq:brief_descriptor}
f_{N_d}(p) := \sum_{i=1}^{N_d} 2^{i-1} \tau(P; x_i, y_i).
\end{equation}

\begin{figure}[htbp]
    \centering
    \begin{minipage}[b]{0.2\textwidth}
        \centering
        \includegraphics[width=\textwidth]{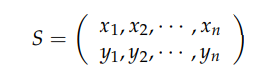}
    \end{minipage}
    \hfill
    \begin{minipage}[b]{0.5\textwidth}
        \centering
        \includegraphics[width=\textwidth]{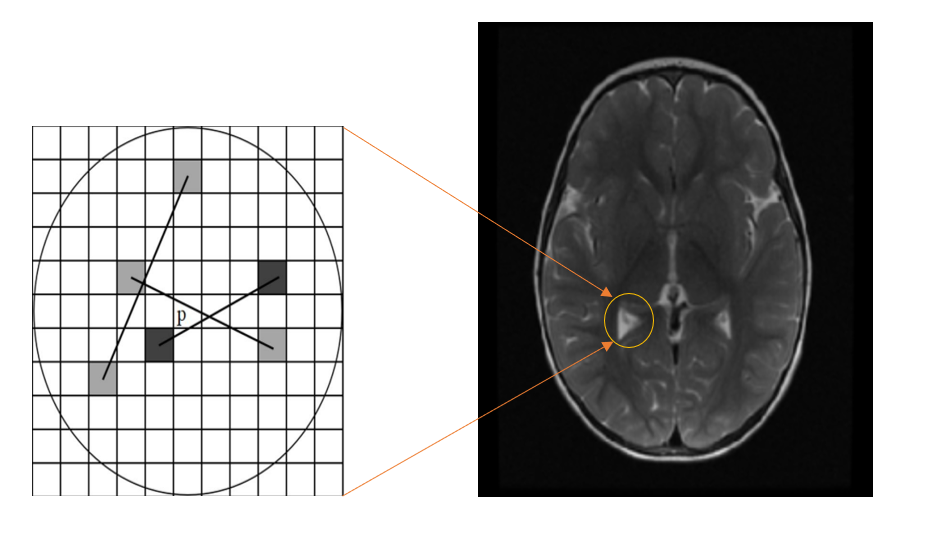}
        \caption{Schematic diagram of the descriptor calculation}
        \label{fig:descriptor_schema}
    \end{minipage}
\end{figure}

The descriptor length can be set to 128, 256, 512 bits, etc., allowing a trade off between speed, storage efficiency, and recognition performance. Since BRIEF considers only single pixels, it is sensitive to noise. To mitigate this, ORB computes each test using a $5 \times 5$ sub-window within a $31 \times 31$ neighborhood. The schematic diagram of this descriptor calculation is illustrated in Fig. \ref{fig:descriptor_schema}.

BRIEF is undirected and lacks rotation invariance. ORB addresses this by assigning an orientation to each descriptor. For any $n$ binary tests at positions $(x_i, y_i)$, a $2 \times n$ matrix $S$ is defined. Using the neighborhood direction $\theta$ and the corresponding rotation matrix $R_{\theta}$, the corrected version of the neighborhood $S_{\theta}$ is given by \eqref{eq:rotation}:

\begin{equation} \label{eq:rotation}
S_{\theta} = R_{\theta} S.
\end{equation}

Thus, the steered BRIEF descriptor becomes:

\begin{equation} \label{eq:steered_brief}
g_n(p, \theta) := f_{N_d}(p) \mid (x_i, y_i) \in S_{\theta}.
\end{equation}

After obtaining the steered BRIEF, a greedy search is performed to select 256 pixel block pairs with the lowest correlation, resulting in the final descriptor~\cite{Xie2022}.

\vspace{2cm}
\subsection{The Flowchart of  ORB Algorithm}\label{SCM}
A flowchart of the ORB algorithm is presented in Fig. \ref{fig:example}:  

\begin{figure}[htbp]
    \centering
    \includegraphics[width=0.18\textwidth]{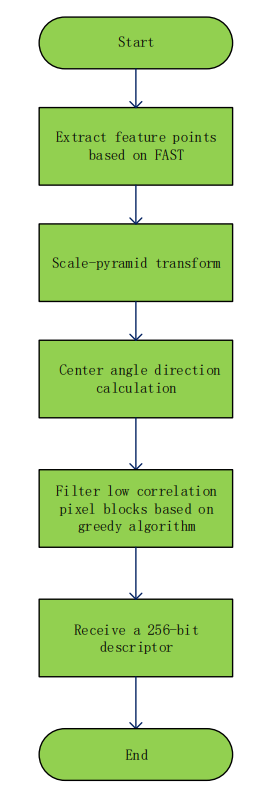}
    \caption{ORB feature extraction and the matching flowchart.}
    \label{fig:example}
\end{figure}

\subsection{Methodology of Bag of Words (BoW)}
The Bag of Words (BoW) model, originally established in Natural Language Processing (NLP), is utilized in this framework for computer vision to represent MRI scans as a collection of discrete "visual words" rather than continuous pixel level data~\cite{Juluru2021}. By prioritizing feature frequency over global spatial arrangements, the BoW approach transforms high dimensional local descriptors into a structured and compact feature vector.

Following the ORB feature extraction phase, an extensive set of descriptors is generated. To manage this high-dimensional data, K-Means clustering is implemented to categorize similar descriptors into $k$ distinct clusters (where $k=150$ in this study). Each cluster centroid ($c_j$) serves as a "visual word" within the constructed vocabulary. The optimization objective is to minimize the Sum of Squared Errors (SSE), as defined in \eqref{eq:sse}:

\begin{equation} \label{eq:sse}
J = \sum_{j=1}^{k} \sum_{i=1}^{n} \|x_i^{(j)} - c_j\|^2
\end{equation}

where $x_i$ represents the local descriptors and $c_j$ denotes the $j$-th cluster centroid.

For each individual image, descriptors are mapped to the nearest visual word using the Euclidean distance metric, as expressed in \eqref{eq:matching}:

\begin{equation} \label{eq:matching}
w = \arg\min_{j} \|f - c_j\|
\end{equation}

The occurrence frequency of each visual word is subsequently aggregated to form the final BoW vector. This methodology achieves a substantial data reduction of approximately 99.5\%, distilling the complex image information into a concise representation of its most salient pathological patterns.

These normalized vectors, processed via a standard scaling pipeline, serve as the input for the Support Vector Machine (SVM) classifier. While the BoW model omits explicit spatial context, its capacity to compress high dimensional raw data into a refined representation provides a robust and resource efficient mechanism for brain tumor detection.

\subsection{Support Vector Machines (SVM)}
In this phase, the normalized BoW vectors comprising 150 dimensional feature sets processed via StandardScaler are utilized as the primary inputs for a Support Vector Machine (SVM) classifier~\cite{Hearst1998}. The SVM is configured to establish an optimal hyper plane that maximizes the geometric margin between classes while simultaneously minimizing the structural risk of classification error.

Given the training dataset $(\mathbf{x}_i, y_i)$, where $\mathbf{x}_i \in \mathbb{R}^{150}$ and $y_i \in \{-1, +1\}$, the SVM optimizes the objective function presented in \eqref{eq:svm_opt}:

\begin{equation} \label{eq:svm_opt}
\min_{\mathbf{w}, b, \boldsymbol{\xi}} \frac{1}{2} \|\mathbf{w}\|^2 + C \sum_{i=1}^{n} \xi_i
\end{equation}

subject to the following constraints:

\begin{equation} \label{eq:svm_constraint}
y_i (\mathbf{w}^T \phi(\mathbf{x}_i) + b) \geq 1 - \xi_i, \quad \xi_i \geq 0
\end{equation}

where the regularization parameter $C$ governs the trade off between margin maximization and the tolerance for misclassification.

To address the non linear distribution within the BoW feature space, a Radial Basis Function (RBF) kernel is employed, as defined in \eqref{eq:rbf_kernel}:

\begin{equation} \label{eq:rbf_kernel}
K(\mathbf{x}_i, \mathbf{x}_j) = \exp(-\gamma \|\mathbf{x}_i - \mathbf{x}_j\|^2)
\end{equation}

\begin{figure}[htbp]
    \centering
    \includegraphics[width=0.48\textwidth]{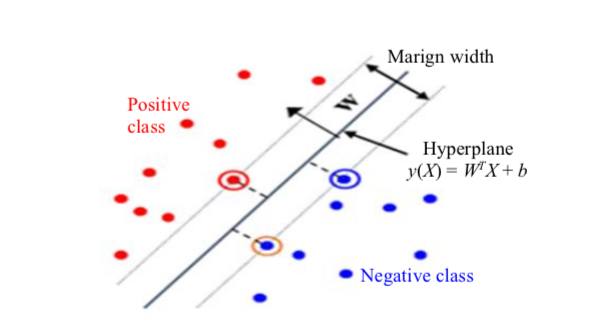}
    \caption{Conceptual representation of a binary classification via a maximum-margin SVM hyper-plane.}
    \label{fig:svm_classification}
\end{figure}

The primary hyperparameters, $C$ and $\gamma$, were utilized with their default configurations ($C = 1.0$, $\gamma = 1/n_{\text{features}}$). Empirically, this configuration yielded the highest classification accuracy of 97.5\% across all evaluated parameter variations. The fundamental principle of this classification mechanism is visually demonstrated in Fig. \ref{fig:svm_classification}.

\subsection{Methodological Framework} 
\begin{figure}[htbp]
    \centering
    \includegraphics[width=0.52\textwidth]{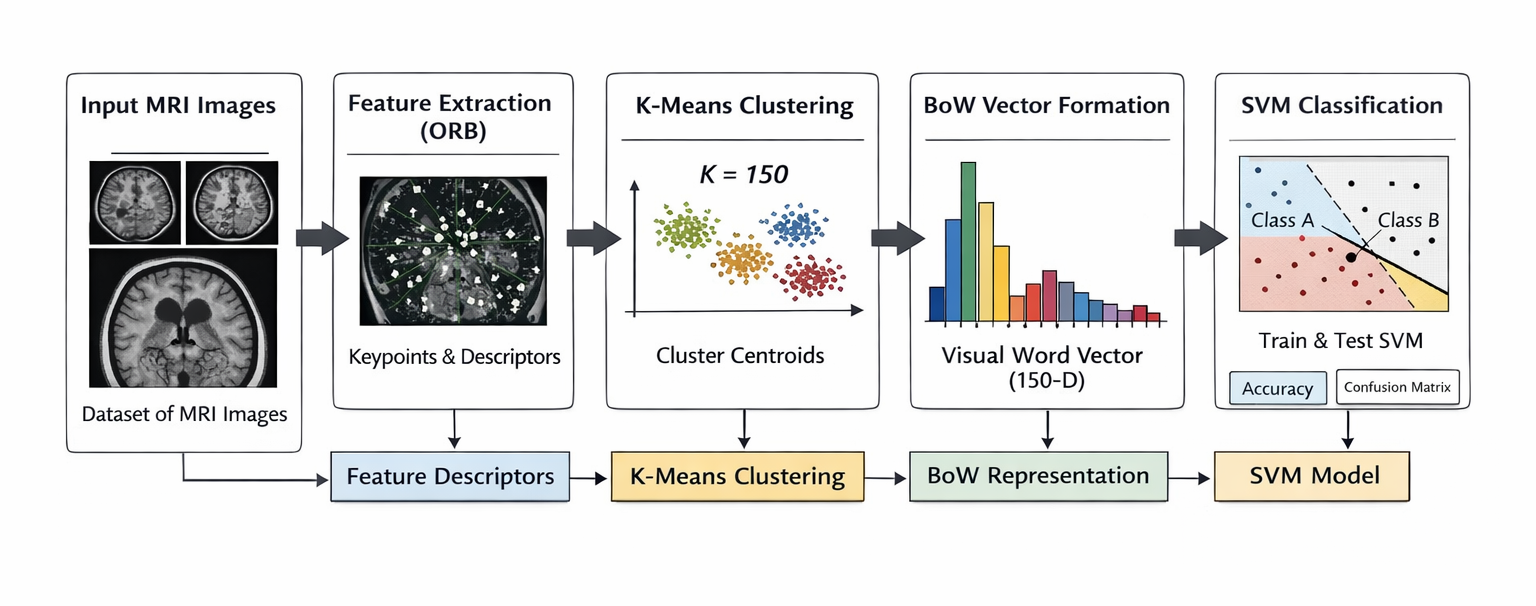}
    \caption{Flowchart depicting the sequential stages of the proposed ORB-SVM diagnostic methodology.}
    \label{fig:methodology_flowchart}
\end{figure}

The end-to-end operational sequence of the proposed diagnostic system, ranging from initial image acquisition to final tumor classification, is systematically illustrated in the flowchart provided in Fig. \ref{fig:methodology_flowchart}.

\section{Proposed Methodology for Brain Tumor Detection}
This section delineates a hybrid diagnostic framework for brain tumor detection from MRI scans, integrating Oriented FAST and Rotated BRIEF (ORB) for feature extraction with a Support Vector Machine (SVM) classifier. The initial stage of the methodology involves the application of the ORB algorithm to identify and describe salient keypoints within each image. These keypoints correspond to critical structural cues and intensity variations that distinguish pathological tissue from healthy anatomical structures.

To enhance computational efficiency while maintaining diagnostic integrity, the high dimensional descriptors generated by ORB are subsequently compressed. This selective dimensionality reduction focuses on preserving essential morphological patterns and filtering out redundant information. This process ensures that the resulting feature set remains highly representative of the underlying pathology, facilitating a more robust classification process.

In the final stage, the refined feature set is utilized to train and evaluate an SVM classifier. The SVM establishes an optimal decision boundary to differentiate between malignant and benign instances based on the extracted visual descriptors. By combining the precision of ORB in capturing localized visual structures with the SVM's capacity for high-dimensional classification, the proposed framework improves both diagnostic accuracy and resource efficiency.

\section{Results}
In this section, we evaluate the performance of the proposed ORB-SVM framework. The model's effectiveness is summarized using a confusion matrix and key performance metrics including Accuracy, Precision, and Sensitivity.

The following metrics, as summarized in Table \ref{tab:performance_metrics}, demonstrate the model's robustness:

\begin{table}[h!]
\centering
\caption{Performance Metrics of the ORB-SVM Framework.}
\label{tab:performance_metrics}
\begin{tabular}{|l|c|c|c|}
\hline
\textbf{Metric} & \textbf{Formula} & \textbf{Calculation} & \textbf{Value} \\
\hline
Accuracy  
& $\frac{TP + TN}{\text{Total}}$ 
& $\frac{291 + 294}{600}$ 
& $97.50\%$ \\
\hline
Precision 
& $\frac{TP}{TP + FP}$ 
& $\frac{291}{300}$ 
& $97.00\%$ \\
\hline
Recall (Sensitivity) 
& $\frac{TP}{TP + FN}$ 
& $\frac{291}{297}$ 
& $\sim 98\%$ \\
\hline
\end{tabular}
\end{table}

The ORB-SVM framework demonstrates high efficiency. By reducing the input data by 99.5\% through ORB feature extraction and BoW encoding, the model achieves a high accuracy of 97.5\% while maintaining a significantly lower computational footprint compared to deep learning architectures.
   
\begin{figure}[h]    
    \centering
    \includegraphics[width=0.25\textwidth]{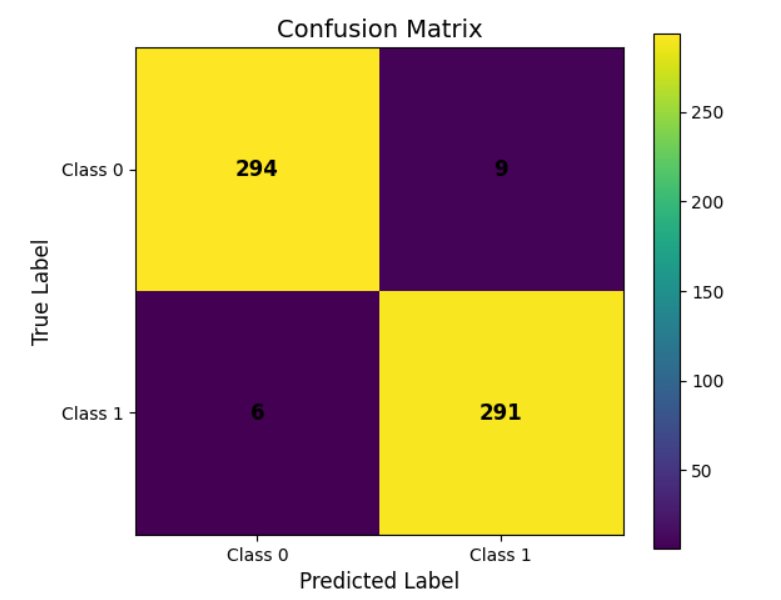}
    \caption{Model's confusion matrix}
    \label{fig:confusion_matrix}
\end{figure}

The confusion matrix shown in Fig. \ref{fig:confusion_matrix} illustrates the distribution of correctly and incorrectly classified instances across the two classes, providing a comprehensive overview of the model's predictive behavior.

\section{Comparison with Other Models}
To evaluate the effectiveness of the proposed ORB-SVM framework, we compare its performance against several established deep learning architectures on the same brain tumor detection task. The experimental results, summarized in Table \ref{tab:comparison}, present a comprehensive comparison of model complexity and classification accuracy.

\begin{table}[h]
\centering
\caption{Comparison of Model Complexity and Classification Accuracy.}
\label{tab:comparison}
\hspace*{-0.8cm}
\begin{tabular}{@{}lccr@{}}
\toprule
\textbf{Model} & \textbf{Architecture} & \textbf{Parameters} & \textbf{Accuracy} \\
\midrule
\textbf{Proposed ORB-SVM} & ORB + BoW + SVM & \textbf{12,349} & \textbf{97.50\%} \\
Custom CNN & 3× Conv2D + 2× Dense & 3,284,159 & $\sim$97.16\% \\
Xception & Depthwise Separable Conv & 20,863,529 & $\sim$98.0\% \\
DenseNet169 & Dense Connections & 12,642,880 & $\sim$96.0\% \\
InceptionResNetV2 & Inception + Residual & 54,338,273 & $\sim$66.0\% \\
ResUNet & Encoder-Decoder + Attention & 31,575,649 & $\sim$93.8\% \\
\bottomrule
\end{tabular}
\end{table}

The experimental results reveal several important findings. The proposed ORB-SVM framework achieves 97.50\% accuracy with merely \textbf{12,349 parameters}, which is approximately \textbf{1,689 times smaller} than Xception (20.86M parameters) and \textbf{4,400 times smaller} than InceptionResNetV2 (54.34M parameters). This represents a \textbf{99.94\% }less parameters while maintaining competitive accuracy.

Notably, InceptionResNetV2, despite containing 54.34M parameters, achieves only 66\% accuracy \textbf{31.5\% lower} than the proposed method. This substantial performance gap suggests that generic deep learning architectures may overfit to noise in medical images, while the ORB feature extraction captures clinically relevant patterns more effectively.

The custom 3 layer CNN achieves slightly lower accuracy (97.16\%) than our method and requires \textbf{266× more parameters}. This trade-off becomes particularly relevant in resource constrained clinical environments where computational efficiency is paramount.

\section{Summary and Conclusions}
This research investigated a novel diagnostic framework for brain tumor detection by integrating Oriented FAST and Rotated BRIEF (ORB) feature extraction with a Support Vector Machine (SVM) classifier. This synergy provides a critical balance between computational efficiency and diagnostic fidelity, enabling the system to differentiate between pathological and healthy tissues without compromising the subtle structural details essential for clinical accuracy.

Experimental evaluations demonstrated that the framework achieves a classification accuracy of 97.5\%. Furthermore, the substantial dimensionality reduction of approximately 99.5\% significantly optimizes the feature space, reducing the risk of overfitting and ensuring the model's robustness. Due to its minimal requirement for high end computational infrastructure, this methodology remains highly feasible for deployment in medical institutions with limited technical resources, offering a scalable alternative to parameter heavy architectures.

Despite these promising results, certain limitations remain. The current study utilized a specific dataset, and the inherent variability in tumor morphology across different stages and types necessitates further investigation. Future work will focus on expanding the diversity of the training datasets to evaluate the framework's generalizability and to test the model on multi class datasets. Additionally, exploring hybrid architectures that integrate ORB descriptors with deep learning components, such as Convolutional Neural Networks (CNNs), may further enhance feature representation and improve diagnostic precision in complex clinical scenarios.

\bibliographystyle{IEEEtran}
\bibliography{references}

\end{document}